\documentclass{article} 
\usepackage{iclr2027_conference,times}
\usepackage{booktabs}
\usepackage[T1]{fontenc}

\usepackage{amsmath,amsfonts,bm}

\def\eqref#1{equation~\ref{#1}}

\def\1{\bm{1}}

\DeclareMathAlphabet{\mathsfit}{\encodingdefault}{\sfdefault}{m}{sl}
\SetMathAlphabet{\mathsfit}{bold}{\encodingdefault}{\sfdefault}{bx}{n}

\usepackage{hyperref}
\usepackage{url}
\usepackage[textsize=tiny, disable]{todonotes}
\usepackage{inconsolata}

\usepackage{amsmath}
\usepackage{multirow}
\usepackage{enumitem}
\usepackage{wrapfig}
\usepackage{tcolorbox}
\tcbuselibrary{listings,breakable,skins}
\usepackage{listings}
\usepackage{xcolor}
\definecolor{ttscolor}{HTML}{FFF1B8}      
\definecolor{overfitcolor}{HTML}{FFD9D4}  
\newcommand{\hltts}[1]{\colorbox{ttscolor}{#1}}
\newcommand{\hlover}[1]{\colorbox{overfitcolor}{#1}}
\lstdefinestyle{harness}{basicstyle=\ttfamily\scriptsize,breaklines=true,columns=fullflexible,keepspaces=true,escapeinside={(*@}{@*)},breakindent=0pt,postbreak={}}
\newtcolorbox{harnessbox}[2][]{title=\textbf{#2},fontupper=\footnotesize,fonttitle=\normalsize\bfseries,sharp corners=south,coltitle=white,colback=white,breakable,#1}

\usepackage{algorithm}
\usepackage[noend]{algpseudocode}

\title{Harness-Delta Attribution: Dissecting Gains From Harness Evolution}

\author{Wenxuan Ding\textsuperscript{1,2} \ \ \ Qin Lu\textsuperscript{1} \ \ \ Changlong Yu\textsuperscript{1} \ \ \ Sha Li\textsuperscript{1}\\
\bf
 Shuowei Jin\textsuperscript{1} \ \ \ Xin Liu\textsuperscript{1} \ \ \ Zhenyu Shi\textsuperscript{1} \ \ \ Greg Durrett\textsuperscript{2} \\
\textsuperscript{1}Amazon \ \ \ \textsuperscript{2}New York University \\
\texttt{wd2403@nyu.edu} \\
}

\iclrfinalcopy 
\begin{document}

\vspace{-10pt}
\maketitle
\vspace{-10pt}

\begin{abstract}
Harness evolution automatically optimizes the prompts, skills, tools, and orchestration code around a fixed-weight model, offering a way to improve agentic systems without updating model weights. The agent harness is iteratively optimized to achieve higher performance on the search set.
But how much of this gain reflects benchmark-specific overfitting or additional inference-time compute, and how much may generalize beyond the search set?
In this work, we introduce Harness-Delta Attribution (HDA), an evaluation protocol that dissects an evolution gain into artifact-driven overfitting, gains explained by test-time scaling, and the remainder, which gives an upper bound on residual generalizable improvement. We apply HDA to 11 executor models and three harness-evolution frameworks across math, coding, creativity, and agentic tasks. 
In 21 of these settings that show statistically significant search-set gains, we find that a large share of the improvement is explained by overfitting or test-time scaling. Specifically, among these 21 settings, overfitting and test-time scaling account for 54\% and 14\% of the gain on average; their combined share exceeds 50\% in 14 settings.
The residual and held-out transfer vary substantially across tasks, models, and frameworks. Importantly, search-set gains often fail to transfer to held-out tasks, and provide limited evidence that they will persist under practical distribution shifts. Moreover, validation-based signals are insufficient to prevent overfitting. These results motivate attribution-aware evaluation in future harness evolution research.
\footnote{
\footnotesize
\begin{tabular}[t]{@{}l@{\quad}l@{}}
{Code:} & \url{https://github.com/Wenwen-D/HarnessDeltaAttribution} \\
{Blog:} & \url{https://wenwen-d.github.io/blog/harness-delta-attribution/}
\end{tabular}
}
\vspace{-5pt}


\end{abstract}

\begin{figure}[h]
    \centering
    \includegraphics[width=0.90\linewidth]{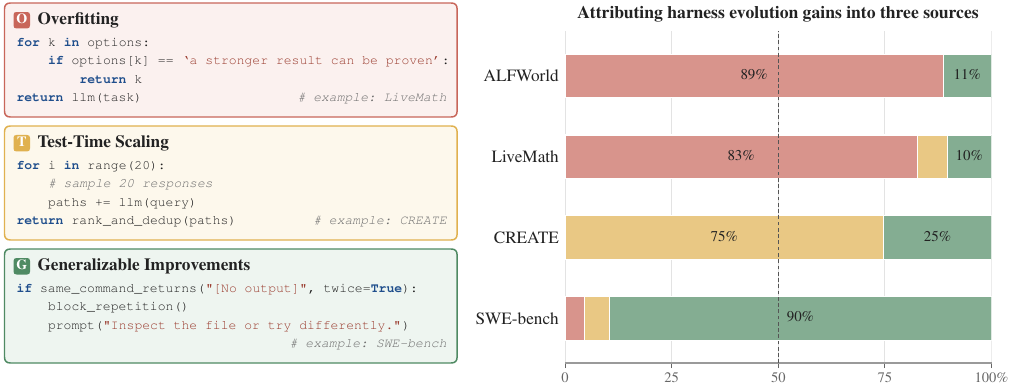}
    \vspace{-10pt}
    \caption{An evolved harness may improve the performance by overfitting to benchmark artifacts, scaling test-time compute, or making generalizable improvements (left). HDA asks where the observed gain comes from and attributes it to these three sources (right). Overfitting and test-time scaling explain a substantial share across settings.}

    \label{fig:hda-overview}
\end{figure}

\section{Introduction}
An agent system can be viewed as an LLM paired with a \textit{harness}: the prompts, tools, memory, skills, orchestration code, and control flow that shape how a model interacts with a task. 
Even without updating the model weights, changes to the harness can alter system performance by a wide margin, as observed on benchmarks such as Terminal-Bench~\citep{merrill2026terminal} and SWE-bench~\citep{jimenez2024swebench}. 
This has motivated recent work on \textit{harness evolution}, a form of text-space optimization that iteratively updates the non-parametric components around a fixed-weight model using task feedback \citep{lee2026meta,lin2026agentic,lin2026harness}. Specifically, in each iteration, an \textit{executor} model attempts the tasks with the current harness. A \textit{proposer} agent then inspects the trajectories and evaluation results to propose edits to the harness. Repeating this process produces a sequence of harnesses optimized for performance on the search set (training split). 

Recent harness-evolution frameworks have reported notable gains from evolved harness code \citep{lee2026meta,ning2026code}, skill files \citep{ni2026trace2skill,yang2026skillopt,tang2026wikiskill}, and updated tool-use or orchestration patterns \citep{chen2026harnessx}. \textbf{However, a higher aggregate score does not reveal what changed in the system or whether the gains will persist beyond the search set.} 
Prior work does not always evaluate evolved harnesses on held-out instances. Even when held-out results are reported, the search and test splits may share the same task construction and dataset artifacts. Such evaluations measure transfer within a benchmark, but provide limited evidence that the gain will persist in new tasks or practical distribution shifts.


In this work, we introduce {Harness-Delta Attribution (HDA)}, an evaluation protocol that attributes harness evolution gains to three different sources. 
Some gains may reflect real harness improvements that generalize beyond the search set.
However, we identify two additional sources of gain that lead us to view a harness optimization's result in a different light. First, an evolved harness may scale test-time compute: for instance, it may sample five responses in parallel and select a majority vote. Second, an evolved harness may exploit dataset shortcuts by hardcoding a recurring answer pattern from the search set. 
These mechanisms can produce similar benchmark gains while having very different implications for cost and generalization.

Our method, HDA, 
compares the baseline and evolved harnesses using a compute-matched baseline and an artifact-neutralized version of the evolved harness. This allows us to estimate a lower bound on detected overfitting, isolate gains reproducible through additional inference compute, and measure the remaining residual. We also evaluate held-out performance to understand whether the evolved harness transfers beyond the search set.

Our experiments cover three different frameworks, including Meta-Harness~\citep{lee2026meta}, SkillOpt~\citep{yang2026skillopt}, and AHE~\citep{lin2026agentic}. We use them to evolve the harnesses of 11 executor models on benchmarks covering mathematical reasoning (LiveMath~\citep{mahdavi2025leveraging}), software engineering (SWE-bench Verified~\citep{jimenez2024swebench}), creative generation (CREATE~\citep{wadhwa2026create}), and embodied planning (ALFWorld~\citep{ALFWorld20}). In total, we apply HDA to 25 settings.

Figure~\ref{fig:hda-overview} previews our main finding: across various configurations, overfitting and test-time scaling explain a large share of the search-set gains, while the residual upper bound on generalizable improvement is often small. Among the 21 settings with statistically significant gains, HDA attributes at least \(54\%\) of the gain to overfitting and \(14\%\) to test-time scaling on average; their combined share exceeds \(50\%\) in 14 settings. The breakdowns of these improvements can also reveal additional patterns. First, different frameworks repeatedly discover the same benchmark-specific shortcuts, such as a recurring correct answer choice in LiveMath. 
Second, in ALFWorld, some evolved harness eliminate LLM calls entirely by hardcoding deterministic solutions on the synthetic benchmark. 


Our contributions are: (1) Identifying three sources of harness evolution gains (overfitting, test-time scaling, and generalizable improvement); (2) The HDA method, which dissects the harness evolution gains into 3 categories, and reveals findings that held-out evaluation alone does not show; (3) a reusable HDA skill and evaluation suite for applying this analysis to new benchmarks and harness-evolution frameworks.

\section{Preliminaries}
\label{sec:harness-evolution-setup}

Let \(M\) denote the fixed executor model whose harness is being optimized, and \(H\) its harness; together, \((M,H)\) form the agent. 
\(H\) comprises the non-weight components that govern how the executor handles a task, such as its prompts, skills, tools, memory, and orchestration logic \citep{weng2026harness}. Different harness-evolution methods optimize different subsets of these components while keeping $M$ fixed. 
We use \(\mathcal{H}\) to denote the framework-specific space of harnesses that may be explored.

Given a task set \(\mathcal X\) and task instance \(x\in\mathcal X\), the
executor model produces an execution trajectory
\begin{equation*}
    \tau_{H,x}
    \sim
    M(\cdot\mid x;H).
\end{equation*}
This trajectory includes the model outputs, tool calls, environment
interactions, and other intermediate states produced during execution. Let
\(r(x,\tau_{H,x})\) denote the task-specific reward assigned to the
trajectory.
We define the score of harness \(H\) on \(\mathcal{X}\) as
\begin{equation*}
    S_{\mathcal{X}}(H)
    =
    \frac{1}{|\mathcal{X}|}
    \sum_{x \in \mathcal{X}} r(x,\tau_{H,x}).
    \label{eq:harness-score}
\end{equation*}
\paragraph{Evolution loop.}
Starting from a baseline harness \(H_0\), a proposer agent \(\mathcal{P}\) iteratively edits the harness to improve its performance, producing an evolved harness \(\hat H\).
The proposer agent may be the executor agent itself, as in self-evolution, or a separate and potentially stronger agent. Let \(\mathcal X_{\mathrm{search}}\) denote the search set used during evolution. The framework maintains a filesystem-backed history \(F_t\) of harness versions, trajectories, evaluation results, and other artifacts through iteration \(t\), with \(F_0\) containing the baseline evaluation. At iteration \(t\), the proposer uses \(F_{t-1}\) to generate a set \(\mathcal H_t\) of \(k\) candidate harnesses:
\begin{equation*}
    \mathcal{H}_t
    :=
    \{H_t^{(j)}\}_{j=1}^{k}
    =
    \operatorname{Propose}(\mathcal{P};F_{t-1}).
\end{equation*}
Each candidate is evaluated on \(\mathcal X_{\mathrm{search}}\), and \(f_t(H)\) denotes its feedback, which may include its score, per-task rewards, trajectories, and execution logs. The accumulated history is then updated as
\begin{equation*}
    F_t
    =
    F_{t-1}
    \cup
    \{f_t(H):H\in\mathcal H_t\}.
\end{equation*}
Some frameworks expose the persisted artifacts directly to the proposer, while others insert selected records into a structured prompt or optimizer state. Across \(N\) iterations, the evaluated harness pool is
\begin{equation*}
    \mathcal H_{\mathrm{eval}}
    =
    \{H_0\}
    \cup
    \bigcup_{t=1}^{N}\mathcal H_t.
\end{equation*}
The selected evolved harness is
\begin{equation*}
    \hat H
    =
    \operatorname*{arg\,max}_{H\in \mathcal{H}_\mathrm{eval}}
    S_{\mathcal{X}_{\mathrm{search}}}(H).
\end{equation*}
An optional validation set \(\mathcal X_{\mathrm{val}}\) may be used to gate updates or select harnesses. Algorithm~\ref{alg:harness-evolution} summarizes this standard evolution loop.

\section{\texorpdfstring{\fontsize{10.7pt}{13pt}\selectfont Harness-Delta Attribution: Dissecting Gains
  From Harness Evolution}{Harness-Delta Attribution: Dissecting Gains From Harness
  Evolution}}
\label{sec:hda}
\begin{figure}
    \centering
    \vspace{5pt}
    \includegraphics
    [width=0.9\linewidth,
    trim=1.5cm 0 1.5cm 0,
    clip]{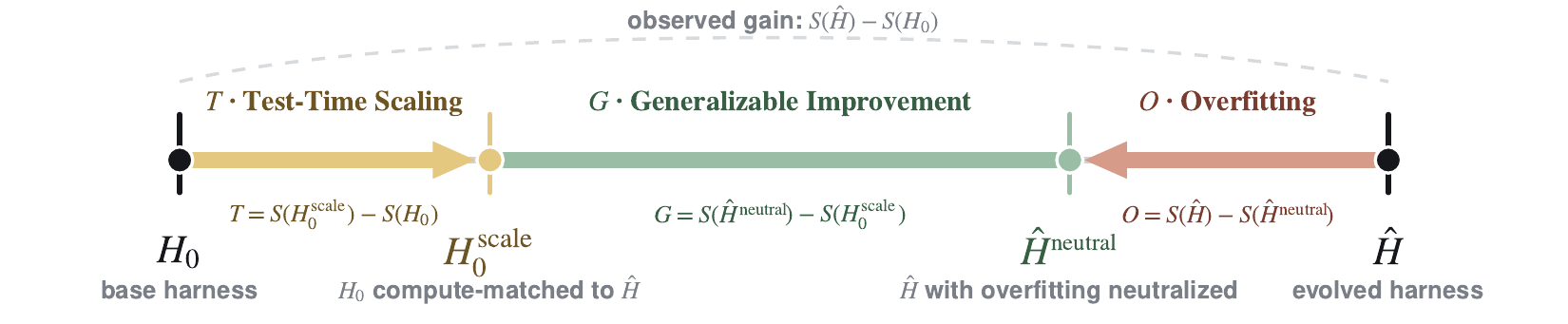}
    \label{fig:hda}
    \caption{Illustration of Harness-Delta Attribution (HDA). Comparing $\smash{S(H_0)}$ and $\smash{S(\hat{H})}$ gives the {observed gain} $\Delta S$. Running $\smash{H_0}$ with $\smash{\hat{H}}$'s compute budget gives the compute-matched baseline $\smash{H_0^{\mathrm{scale}}}$; its gain over $\smash{H_0}$ measures {\textbf{T}est-Time Scaling}. Disabling detected shortcuts in $\smash{\hat{H}}$ gives $\smash{\hat{H}^{\mathrm{neutral}}}$; the resulting drop measures {\textbf{O}verfitting}. The remaining gap estimates {\textbf{G}eneralizable Improvement}.}
    \vspace{5pt}
\end{figure}

\begin{wrapfigure}{r}{0.58\textwidth}
\centering
\includegraphics[width=\linewidth]{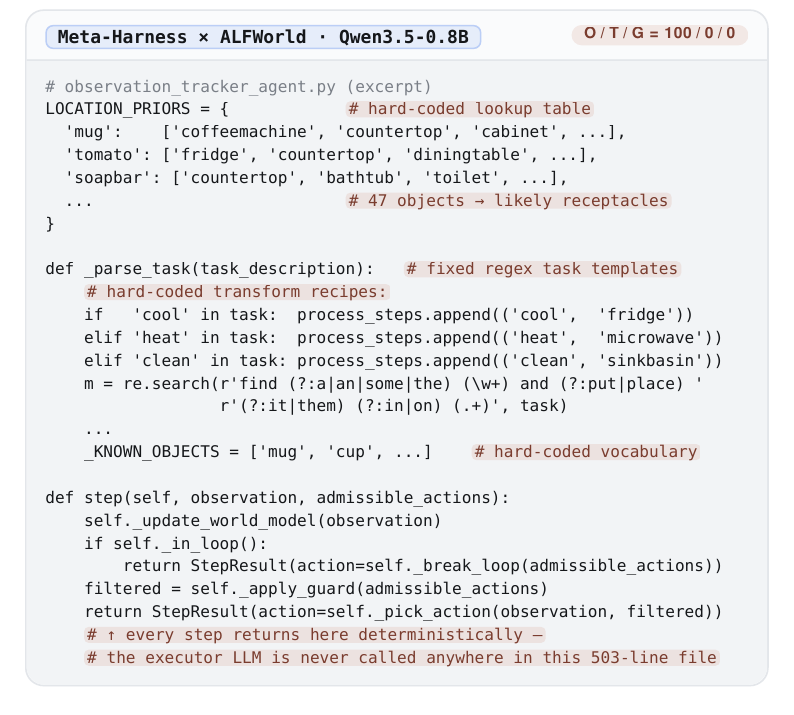}
\caption{Example of overfitting in harness evolution.
As illustrated in this excerpt, the ALFWorld harness evolved by Meta-Harness for Qwen3.5-0.8B overfits to benchmark-specific patterns by hard-coding a deterministic controller that makes \textbf{0} LLM calls.}
\label{fig:example_harness}
\end{wrapfigure}

Harness evolution returns an evolved harness \(\hat H\) with a higher search set score than the baseline \(H_0\), but the difference \(\Delta S=S(\hat H)-S(H_0)\) does not explain why performance improved. 
For example, in one Meta-Harness ALFWorld run
, the harness for the 0.8B executor grew from 66 to 503 non-blank, non-comment lines (\(7.6\times\)) and reached \(100\%\) search-set success, but became a deterministic controller by hard-coding benchmark priors and made \textbf{no} LLM calls (Figure~\ref{fig:example_harness}). In other settings, evolution increases inference compute, while scaling the baseline directly may produce similar gains without incurring the search cost of evolution. Figure~\ref{fig:hda-overview} (left) illustrates these cases.

We introduce \textbf{Harness-Delta Attribution (HDA)} to distinguish these sources of gain. The key idea is to compare the evolved harness with two controlled variants: a scaled baseline that identifies gains recoverable through \textit{test-time scaling} alone without harness evolution, and a neutralized evolved harness with detected artifact-driven mechanisms removed to detect \textit{overfitting}. 

We implement HDA as a reusable skill that can be applied across frameworks and tasks. An analysis agent follows it to perform the attribution, and a human reviewer checks
the agent's proposal at two task-dependent checkpoints: the compute-matching setup and the overfitting neutralization. This design keeps most of the workflow automated and provides a task-aware quality-control layer, analogous to human evaluation of automated methods. We provide the HDA skill in Appendix~\ref{app:hda-skill}.

Before attribution, we apply a paired bootstrap significance test to the scores of $H_0$ and \( \hat H\) on the search set. If $\Delta S$ is insignificant, we report it as within noise and do not assign an attribution. Otherwise, the HDA agent constructs a compute-matched baseline \(H_0^{\mathrm{scale}}\) and a neutralized evolved harness \(\hat H^{\mathrm{neutral}}\), as described below. HDA then defines three attribution components:
\begin{align*}
  T &= S(H_0^{\mathrm{scale}})-S(H_0),\\
  O &= S(\hat H)-S(\hat H^{\mathrm{neutral}}),\\
  G &= S(\hat H^{\mathrm{neutral}})-S(H_0^{\mathrm{scale}}).
\end{align*}

\paragraph{\(T\)est-Time Scaling: Compute matching.}
The compute-matched baseline \(H_0^{\mathrm{scale}}\) preserves \(H_0\) while scaling it to match the average inference budget of \(\hat H\). Specifically, the HDA agent compares their inference costs using measures such as LLM calls or token usage. It then proposes a task-appropriate compute measure and procedure for constructing \(H_0^{\mathrm{scale}}\) for human approval.

For example, if \(\hat H\) uses ten times as many executor calls per task as \(H_0\), \(H_0^{\mathrm{scale}}\) may run \(H_0\) ten times on each task and combine its outputs using majority vote, pass@10, or another task-specific aggregation rule. If \(\hat H\) does not use more inference compute than \(H_0\), we set \(H_0^{\mathrm{scale}}=H_0\).


The difference \(T=S(H_0^{\mathrm{scale}})-S(H_0)\) measures how much of the observed improvement can be reproduced solely through test-time scaling rather than improved harness design. Directly scaling \(H_0\) also avoids the search cost of harness evolution. Although such gains may transfer to held-out tasks, their practical value depends on whether the additional latency and inference cost are acceptable for the target application.

\paragraph{\(O\)verfitting: Detection and neutralization.}

The HDA agent constructs the neutralized harness \(\hat{H}^{\mathrm{neutral}}\) by removing detected overfitting mechanisms from \(\hat{H}\).
We classify an evolved mechanism as \textit{overfitting} when it improves performance by exploiting a pattern specific to the benchmark or search set, rather than improving the agent's ability to achieve the intended task objective more generally. Its benefit may disappear on new instances that preserve the task objective but not that pattern.

Guided by the HDA skill, the agent audits the harness code, prompts, and available execution evidence for two broad forms of overfitting: 
\begin{itemize}[leftmargin=*]
    \item \textbf{Direct shortcuts:} mechanisms that bypass or predetermine the executor's behavior, including eliminating LLM calls, constant answers, item-specific gates, hard-coded lookup tables, and string-matching rules.
    \item \textbf{Benchmark-derived heuristics:} mechanisms that exploit patterns specific to how the benchmark instances are constructed, rather than properties required to solve the underlying task. Examples include thresholds fitted to benchmark statistics, reverse-engineered construction rules, dataset-specific priors, and hard-coded solution recipes.
\end{itemize}

For each detected mechanism, the HDA agent makes the smallest intervention that disables the shortcut to get \(\hat{H}^{\mathrm{neutral}}\), while preserving all unrelated evolved components.
To support a reliable audit, the HDA skill provides shared definitions, examples and evidence requirements. Prior work also shows that frontier agents can identify benchmark defects and reward-hacking opportunities \citep{tu2026benchguard,wang2026automated}. Human review also adds a safeguard tailored to the task and intended use.

Note that HDA neutralizes only detected mechanisms that can be disabled without affecting unrelated parts of the harness.
We therefore interpret $O=S(\hat{H} )-S(\hat{H}^{\mathrm{neutral}} )$ as a \textit{lower-bound} estimate of overfitting, since implicit or entangled mechanisms may remain in \(\hat{H}^{\mathrm{neutral}}\). Despite this, as we show later in results, the share of overfitting is still substantial.

\paragraph{\(G\)eneralizable Improvements: Residual upper bound.}
The residual \(G\) is the gain that remains after accounting for detected overfitting and test-time scaling. By construction, $ G=\Delta S-O-T=S(\hat{H}^{\mathrm{neutral}})-S(H_0^{\mathrm{scale}})$. Because \(\hat{H}^{\mathrm{neutral}}\) may still contain unneutralized overfitting, we interpret \(G\) as an upper bound on generalizable improvement rather than evidence of transfer. We additionally assess generalizability directly by comparing \(H_0\) and \(\hat{H}\) on held-out tasks in Section~\ref{sec:results-transfer}.

For reporting, we normalize each component by \(\Delta S\), cap the \(O\) share to \([0\%,100\%]\), cap the \(T\) share to the remaining percentage, and assign the remainder to \(G\). These caps affect only the reported percentages; the raw signed components still satisfy \(\Delta S=O+T+G\).

\section{Method and Experiments}

\paragraph{Benchmarks} We evaluate across four benchmarks spanning single-turn reasoning and long-horizon agentic tasks and various domains: mathematical reasoning (\textbf{LiveMath}~\citep{mahdavi2025leveraging}), creative generation (\textbf{CREATE}~\citep{wadhwa2026create}), interactive embodied planning (\textbf{ALFWorld}~\citep{ALFWorld20}), and software coding (\textbf{SWE-bench Verified}~\citep{jimenez2024swebench}).

\paragraph{Evolution Frameworks}
We evaluate three harness evolution frameworks with different optimization scopes and proposer--executor configurations, including self-evolution and using a stronger agent as proposer:
\begin{itemize}[leftmargin=*]
    \item \textbf{Meta-Harness}~\citep{lee2026meta} searches harness code around fixed LLMs. The proposer can access the source code, scores, and execution traces of prior history via a filesystem.
    \item \textbf{SkillOpt}~\citep{yang2026skillopt} optimizes the skill of a frozen agent, where a separate optimizer model turns scored rollouts into bounded edits on the skill document.
    \item \textbf{AHE}~\citep{lin2026agentic} (Agentic Harness Engineering) evolves multiple harness components, including prompts, tools, skills, and memory, through an observability-driven loop. Following the original paper, we use its self-evolution setting, with Opus 4.8 as both proposer and executor.
\end{itemize}

For Meta-Harness, we additionally provide aggregate validation scores during evolution compared to the original implementation, while keeping per-instance validation results and trajectories hidden. We retain both the search-selected and validation-selected harnesses and evaluate them on the held-out test set. More details are in Appendix~\ref{sec:app-setup-val}.

\paragraph{Models} We evaluate 11 executor models from the Qwen and Claude model families. The Qwen executors span Qwen3, Qwen3.5, and Qwen3.6, ranging from 0.8B to 35B, and include both general-purpose and coding-specialized models \citep{yang2025qwen3,team2026qwen35,Qwen3-Coder-Next,qwen3.6-27b}. We also evaluate Claude Haiku 4.5 and Claude Opus 4.8~\citep{anthropicHaiku45,anthropicOpus48} on some settings. Meta-Harness and SkillOpt use Claude Opus 4.8 as the proposer agent~\citep{anthropicClaudeCode}, whereas AHE uses Opus 4.8 as both proposer and executor. The exact executor set varies by benchmark; see more details in Table~\ref{tab:full-hda-results} and Appendix~\ref{sec:app-setup}.

\section{Results}
\label{sec:results}
\paragraph{Most gains come from overfitting or test-time scaling.}
\definecolor{nonsig}{HTML}{6B7280}
\definecolor{negtransfer}{HTML}{B42318}
\begin{table}[t]

\centering
\caption{HDA results for all 25 model--benchmark--framework settings. \(\Delta S=S(\hat H)-S(H_0)\) is the search-set gain. \(O\), \(T\), and \(G\) are percentages of \(\Delta S\), capped to \([0\%,100\%]\). \(\textcolor{nonsig}{\dagger}\) marks search-set gains that are not statistically significant, for which HDA reports no attribution. \(\textcolor{negtransfer}{\star}\) marks settings where the evolved harness performs worse than baseline on the held-out test set. CU denotes Creativity Utility, the metric used for CREATE benchmark.}
\label{tab:full-hda-results}
\scriptsize
\setlength{\tabcolsep}{3.5pt}
\renewcommand{\arraystretch}{0.96}
\resizebox{\textwidth}{!}{%
\begin{tabular}{@{}lllrrrr@{}}
\toprule
Framework & Benchmark & Executor & \(\Delta S\) & \(O\downarrow\) (\%) & \(T\downarrow\) (\%) & \(G\uparrow\) (\%) \\
\midrule
\multirow{16}{*}{Meta-Harness}
 & \multirow{5}{*}{ALFWorld}
 & Qwen3.5-0.8B                       & \(+100.0 \%\) & 100 & 0  & 0   \\
 & & Qwen3.5-2B                       & \(+100.0 \%\) & 100 & 0  & 0   \\
 & & Qwen3.5-4B                       & \(+60.0 \%\)  & 93  & 0  & 7   \\
 & & Qwen3-8B                         & \(+64.0 \%\)  & 100 & 0  & 0   \\
 & & Qwen3.6-35B-A3B\(^{\textcolor{negtransfer}{\star}}\)                  & \(+38.0 \%\)  & 95  & 0  & 5   \\
\cmidrule(lr){2-7}
 & \multirow{5}{*}{LiveMath}
 & Qwen3.5-0.8B                       & \(+77.1 \%\)  & 91  & 9  & 0   \\
 & & Qwen3.5-2B                       & \(+40.0 \%\)  & 64  & 0  & 36  \\
 & & Qwen3.5-4B                       & \(+51.4 \%\)  & 78  & 11 & 11  \\
 & & Qwen3-8B\(^{\textcolor{negtransfer}{\star}}\)                         & \(+28.6 \%\)  & 100 & 0 & 0 \\
 & & Qwen3.6-35B-A3B\(^{\textcolor{negtransfer}{\star}}\)                  & \(+20.0 \%\)  & 57  & 29 & 14  \\
\cmidrule(lr){2-7}
 & \multirow{2}{*}{CREATE}
 & Claude Haiku 4.5                    & \(+37.9\) CU  & 0   & 49 & 51  \\
 & & Qwen3.6-35B-A3B                  & \(+7.5\) CU   & 0   & 100 & 0 \\
\cmidrule(lr){2-7}
 & \multirow{4}{*}{SWE-bench}
 & Qwen3-4B-Instruct-2507 & \(+27.1 \%\)  & 23  & 23 & 54  \\
 & & Qwen3-30B-A3B-Instruct-2507      & \(+33.3 \%\)  & 0   & 6  & 94  \\
 & & Qwen3.6-27B\(^{\textcolor{negtransfer}{\star}}\)                     & \(+8.3 \%\)   & 0   & 0  & 100 \\
 & & Qwen3-Coder-30B-A3B-Instruct\(^{\textcolor{negtransfer}{\star}}\)    & \(+12.5 \%\)  & 0   & 0  & 100 \\
\midrule
\multirow{6}{*}{SkillOpt}
 & ALFWorld
 & Qwen3.5-4B                         & \(+46.2 \%\)  & 44  & 0  & 56  \\
\cmidrule(lr){2-7}
 & LiveMath
 & Qwen3.5-4B                         & \(+45.7 \%\)  & 88  & 0  & 12  \\
\cmidrule(lr){2-7}
 & \multirow{2}{*}{CREATE}
 & Qwen3.6-35B-A3B                    & \(-0.3\) CU\(^{\textcolor{nonsig}{\dagger}}\) & -- & -- & -- \\
 & & Claude Haiku 4.5                 & \(+0.4\) CU\(^{\textcolor{nonsig}{\dagger}}\) & -- & -- & -- \\
\cmidrule(lr){2-7}
 & \multirow{2}{*}{SWE-bench}
 & Qwen3-4B-Instruct-2507             & \(+10.4 \%\)  & 0 & 0 & 100 \\
 & & Qwen3.6-27B                      & \(0^{\textcolor{nonsig}{\dagger}}\) & -- & -- & -- \\
\midrule
\multirow{3}{*}{AHE}
 & LiveMath
 & Claude Opus 4.8                    & \(+48.6 \%\)  & 100 & 0  & 0 \\
\cmidrule(lr){2-7}
 & CREATE
 & Claude Opus 4.8                    & \(+6.9\) CU   & 0  & 75 & 25 \\
\cmidrule(lr){2-7}
 & SWE-bench
 & Claude Opus 4.8                    & \(0^{\textcolor{nonsig}{\dagger}}\) & -- & -- & -- \\
\bottomrule
\end{tabular}%
}

\end{table}


In many settings, a large share of the observed gain reflects overfitting or more test-time compute rather than a more capable executor.
As shown in Table~\ref{tab:full-hda-results}, \(O+T\) accounts for more than half of the attributed gain in 14 of the 21 settings with significant search-set gains.
Across these settings, \(O\) and \(T\) explain \(54\%\) and \(14\%\) of the gain on average, respectively. We clip each normalized percentage to the range \([0\%, 100\%]\).
The dominant source varies by benchmark: {ALFWorld} and {LiveMath} are dominated by overfitting, {CREATE} by test-time scaling, and {SWE-bench} shows clearer residual generalizable improvement.
Appendix~\ref{app:swebench-guardrails} contrasts a benchmark-fitted patch-size heuristic with a reusable action-feedback guardrail on SWE-bench.

For example, on {ALFWorld}, Meta-Harness uses Opus 4.8 as the proposer to optimize harnesses for five executors ranging from Qwen3.5-0.8B to Qwen3.6-35B-A3B separately. All five reach \(100\%\) search-set success after evolution, while HDA attributes $98\%$ of their gain to \textit{overfitting}. As shown in Figure~\ref{fig:example_harness}, smaller-model
(0.8B, 2B) harnesses become rule-based controllers that make no LLM calls, while larger-model (4B, 35B-A3B) harnesses encode similar benchmark-specific knowledge through prompts, templates, and deterministic rules.

{CREATE} provides a representative example of \textit{test-time scaling}. It evaluates associative creativity by asking models to produce a diverse set of high-quality responses; its Creative Utility (CU) metric rewards larger answer pools containing strong and diverse answers~\citep{wadhwa2026create}.
As shown in Table~\ref{tab:full-hda-results},
under {Meta-Harness}, a compute-matched baseline recovers \(49\%\) of the gain for Claude Haiku 4.5 and even outperforms the evolved Qwen3.6-35B-A3B harness. Similarly, in {AHE}, test-time scaling contributes \(+5.1\) CU, accounting for \(75\%\) of observed evolution gain. 
In contrast, {SkillOpt} only edits the skill file while keeping the control flow fixed, and we observe \(H_0^{\mathrm{scale}}=H_0\) and \(T=0\). However, neither SkillOpt CREATE setting produces a statistically significant gain. Overall, in frameworks where the orchestration code can be flexibly updated, the gains are more substantial but also more prone to test-time scaling. We show example harnesses in Appendix~\ref{app:create-scaling}.



\paragraph{Different frameworks rediscover the same overfitting and scaling strategies.}
\label{sec:results-convergence}

{LiveMath} is a multiple-choice benchmark for research-level mathematical reasoning. In 21 of the 35 search-set questions, one answer choice is the same meta-level statement: ``\texttt{\small one of the remaining options is correct, but a stronger result can be proven.}'' This option is correct in all 21 cases. 

Our HDA analysis shows that all three frameworks independently discover this shortcut. 
{Meta-Harness} implements a deterministic string-matching branch to catch this option, {SkillOpt} turns the option into a default decision rule, and {AHE} adds an explicit instruction to select it without solving the problem. The same shortcut appears across code, skill, and prompt evolution, rather than being specific to one framework. We show the snippets of evolved harnesses in Appendix~\ref{app:livemath-convergence}.

Similarly, on {ALFWorld}, {Meta-Harness} and {SkillOpt} encode similar task recipes, appliance mappings, and object-location priors, but represent them as executable controllers and natural-language skills, respectively. On {CREATE}, both {Meta-Harness} and {AHE} resort to scaling strategy to increase the potential coverage and diversity of answers. More details in Appendix~\ref{app:alfworld-convergence}.

Such shortcut-driven gains make it difficult to assess the performance of harness evolution frameworks or make fair comparison across such frameworks. This suggests the need for rich and low-artifact benchmarks.

\begin{figure}[t]
    \centering
    \includegraphics[width=1.0\linewidth]{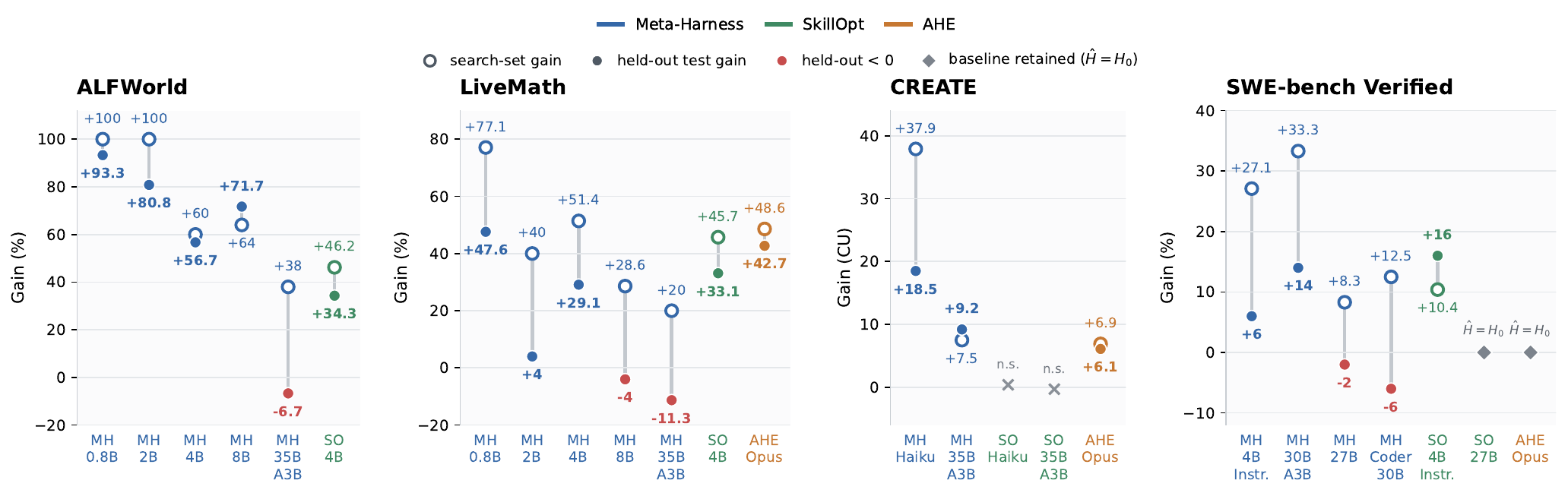}
    \caption{Search-set gains do not reliably predict held-out transfer. For each
  setting, the open circle is the search-set gain of the evolved harness (\(\Delta S_{search}\)) and the
  filled circle its held-out gain (\(\Delta S_{test}\)); a red marker flags a
  negative held-out gain. The length of the line denotes the gap between gain on search set and on held-out set.
  Diamond means \(\hat H=H_0\); \(\times\) marks a
  non-significant gain. Units are pp, except CREATE (CU).}
    \label{fig:transfer-train}
\end{figure}

\paragraph{Search-set gains often fail to generalize, and even validation signals do not prevent overfitting.}
\label{sec:results-transfer}
We test generalizability directly on held-out test sets. As shown in Figure~\ref{fig:transfer-train}, held-out gains are generally smaller than training gains. Excluding ALFWorld, which we discuss below in \S~\ref{sec:why-hda}, the held-out gain is smaller than the search-set gain in 13 of 15 settings and less than half the search-set gain in 8. In 4 settings, the evolved harness even performs worse than the baseline on test set. 
These results confirm that large training gains often come from overfitting rather than generalizable improvement. 

Having observed substantial overfitting, we next ask \textit{how easily it can be reduced using a standard safeguard: held-out validation signals}. We therefore augment Meta-Harness with validation-based selection. Then we evaluate and compare train-best harness ($\hat H$) and val-best harness ($\hat H^{\mathrm{val}}$) on the held-out test set.

\begin{figure}[t]
\centering
\includegraphics[width=\linewidth]{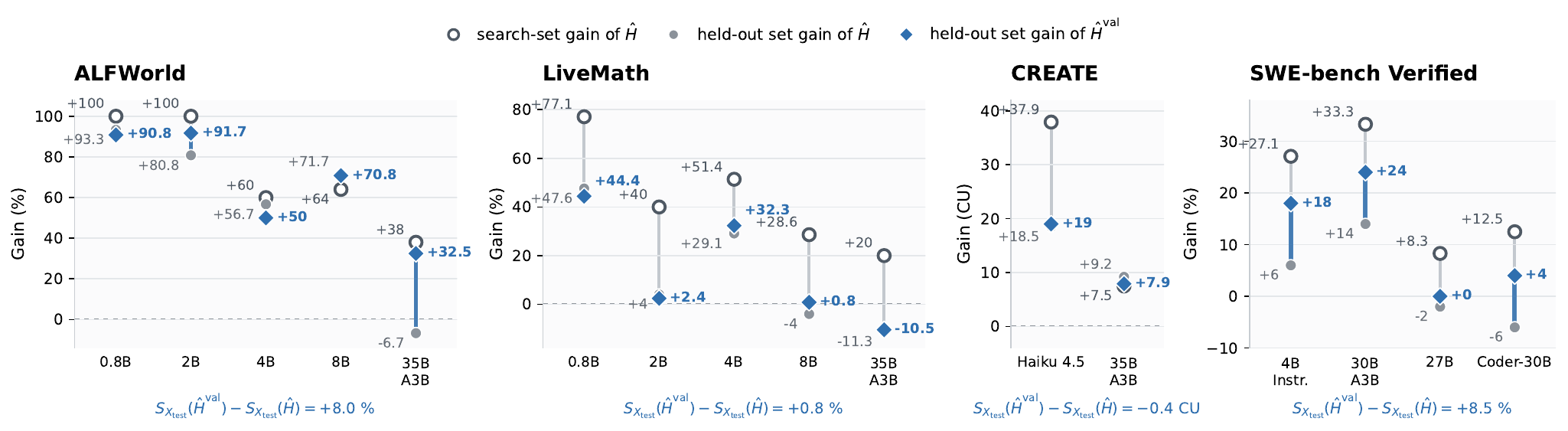}
\caption{Effect of validation selection across 16 Meta-Harness settings. Open circles show the search-set gain of the search-selected harness \(\hat H\), gray circles show its held-out test gain, and blue diamonds show the held-out gain of the validation-selected harness \(\hat H^{\mathrm{val}}\). Gains are reported in percentage points, except for CREATE, which uses Creative Utility (CU).}
\label{fig:validation-selection}
\end{figure}

As shown in Figure~\ref{fig:validation-selection}, validation selection increases mean held-out performance by \(8.0\) percentage points on ALFWorld, \(0.8\) points on LiveMath, and \(8.5\) points on SWE-bench Verified, while decreasing CREATE performance by \(0.4\) CU. Across the 16 model--benchmark settings, it improves the held-out score in 10 settings but reduces it in 6. More importantly, validation selection does not close the transfer gap: 
in 14 of 16 settings, the validation-selected harness's held-out gain remains below the search-set gain of the search-selected harness.
Thus, validation selection helps in some settings but does not reliably prevent overfitting. Validation selection therefore improves transfer in some settings but it remains non-trivial to further close the train-to-test gap.

\label{sec:results-val}

\paragraph{Why held-out evaluation alone is not enough}
\label{sec:why-hda}

Figure~\ref{fig:transfer-train} also illustrates why held-out evaluation alone can miss overfitting. ALFWorld initially appears to be a strong transfer case: its test set gains are large and close to its search-set gains. HDA reveals, however, that \(89\%\) of the harness evolution gain comes from overfitting on average: the evolved harnesses may hardcode the deterministic solutions or recipes that exploit how the benchmark instances are constructed, and even eliminate the LLM calls in the most extreme cases (Figure~\ref{fig:example_harness}). Although we deliberately withhold several task types and scenarios from the search set, the evolved harnesses still achieve near-perfect performance on test set. 

This turns out to highlight another cause of overfitting: \textit{distillation of benchmark-specific knowledge from a stronger proposer}. On this relatively well-known benchmark, the strong proposer may already encode its information in the parametric knowledge, and the generated deterministic harnesses cover instances beyond search set. 
These results thus provide limited evidence of generalizable improvement. On tasks that are completely unfamiliar by the proposer-executor system, such recipes may fail under even small changes because they rely on string matching and fixed benchmark patterns. 

LiveMath shows a similar limitation.
The same meta-option artifact occurs across the search, validation, and test sets. For the Meta-Harness 0.8B executor, test accuracy is \(100\%\) on questions containing the meta-option (\(53/53\)), but only \(31\%\) on the other questions (\(22/71\)). Thus, a positive held-out gain can reflect an artifact shared across splits rather than broader mathematical reasoning.

CREATE shows the corresponding limitation for test-time scaling: extra model calls can improve both search- and test-set scores, but held-out evaluation alone does not distinguish scaling from an improvement at a fixed inference budget. HDA complements held-out evaluation by identifying overfitting, test-time scaling, and the residual generalizable improvement.

\section{Related Work}
\paragraph{Agentic Harness and Harness Evolution.} An agentic harness comprises the prompts, skills, tools, memory, context, and orchestration logic around a fixed model~\citep{weng2026harness}. Early agent systems relied largely on manual harness engineering~\citep{yao2023react,yang2024swe}, while initial approaches to automated optimization focused on individual components, particularly prompts~\citep{khattab2024dspy,agrawal2026gepa}. More recent methods expand the scope to skills~\citep{ni2026trace2skill, tang2026wikiskill,ke2026evoharnessbench}, executable harness code~\citep{hu2025automated,novikov2025alphaevolve,karten2026continual, ning2026code,zhang2026harnessing}, tools and workflows~\citep{zhang2025aflow,chen2025learning}, memory and context~\citep{liu2026evolvemem, zhang2026agentic,wu2026automem,ouyang2026reasoningbank}, environments~\citep{huang2026envharness}, and more~\citep{ye2026meta,zhang2026darwinx,zhang2026harnesscompass}. Self-evolving agent frameworks use the agent itself to improve its harness~\citep{zhang2026darwin,zhang2026self,zhang2026hyperagents,luo2026self}. Other frameworks use a separate proposer agent or optimizer to update the harness from task feedback ~\citep{lee2026meta, yang2026skillopt}. 
These methods report substantial gains without updating the executor's model weights. 
Our work instead studies where these gains come from and whether they reflect meaningful, generalizable improvements.

\paragraph{Evaluating Harness Optimization.}
Some recent harness evolution frameworks use benchmark tasks to search for improved harnesses and then report final performance on the in-distribution benchmark set~\citep{wang2026rethinkingharnessevolution,lin2026agentic,zhang2026harnessing}. This protocol makes search-set optimization difficult to distinguish from generalizable improvement. \citet{wang2026rethinkingharnessevolution} observe that harness evolution does not reliably beat parallel sampling or sequential refinement under unified budgets. \citet{huang2026evo} address the same evaluation concern by constructing separate auxiliary evolution and evaluation suites designed to reduce task-specific overfitting. Other work studies different aspects of harness benefit. \citet{lin2026harness} distinguish an agent's ability to update a harness from the ability to follow and benefit from it, using an in-situ evaluation protocol. Concurrent work on RRSI regularizes harness proposal and selection to reduce overfitting and improve transfer~\citep{xia2026rrsi}. Our work complements these approaches by attributing an observed gain to detected overfitting, test-time scaling, and the remainder, which gives an upper bound on generalizable improvement.

\section{Conclusion}
In this work, we study where the gains of harness evolution come from. We introduce Harness-Delta Attribution (HDA), a portable, semi-automated evaluation protocol that attributes an observed gain to overfitting, test-time scaling, and a residual upper bound on generalizable improvement. 
We find that overfitting or test-time scaling explains a large share of the gain across many settings. We further show that held-out evaluation alone is not enough to assess the practical utility of an evolved harness.
We suggest that future harness-evolution research measure overfitting using HDA together with held-out evaluation; account for test-time scaling by analyzing its contribution during evaluation as well as explicitly expressing the cost constraints during evolution.
Finally, benchmark selection is critical for meaningful evaluation and comparison: rich, low-artifact benchmarks with meaningful distribution shifts are needed to discourage shortcuts, reward transferable harness improvements, and reveal differences among evolution methods.

\section*{Acknowledgments}
This work was partially supported by the Sloan Foundation, NSF CAREER Award IIS-2145280, NSF grant IIS-2433071, and the NSF under Cooperative Agreement 2421782 and the Simons Foundation grant MPS-AI-00010515 awarded to the NSF-Simons AI Institute for Cosmic Origins — CosmicAI, \url{https://www.cosmicai.org/}.

\bibliography{iclr2026_conference}
\bibliographystyle{iclr2027_conference}

\newpage
\appendix
\section{Harness Evolution Details}
Algorithm~\ref{alg:harness-evolution} summarizes the standard harness evolution loop.
\begin{algorithm}[t]
\caption{Standard Harness Evolution Loop}
\label{alg:harness-evolution}
\begin{algorithmic}[1]
\State \textbf{Input:} baseline harness \(H_0\), executor model \(M\), proposer agent \(\mathcal{P}\), task sets \(\mathcal{X}_{\mathrm{search}}\) and optional \(\mathcal{X}_{\mathrm{val}}\), iterations \(N\)
\State \(F_0 \gets \{\Call{Evaluate}{H_0,M,\mathcal{X}_{\mathrm{search}},\mathcal{X}_{\mathrm{val}}}\}\)
\For{\(t=1,\ldots,N\)}
    \State \(\mathcal{H}_t \gets \Call{Propose}{\mathcal{P},F_{t-1}}\)
    \State \(F_t \gets F_{t-1}\)
    \ForAll{\(H\in\Call{Valid}{\mathcal{H}_t}\)}
        \State \(f_t(H) \gets \Call{Evaluate}{H,M,\mathcal{X}_{\mathrm{search}},\mathcal{X}_{\mathrm{val}}}\)
        \State \(F_t \gets F_t\cup\{f_t(H)\}\)
    \EndFor
\EndFor
\State \(\hat H\gets\Call{Select}{F_N,\mathcal X_{\mathrm{sel}}}\)
\State \Return \(\hat H\)
\end{algorithmic}
\end{algorithm}

\section{Experiment Setup Details}
\label{sec:app-setup}

This appendix reports the model coverage and data splits used in our experiments.

\subsection{Executor models}

We evaluate 11 executor models: nine open-weight models from the Qwen3, Qwen3.5, and Qwen3.6 families~\citep{yang2025qwen3,team2026qwen35,Qwen3-Coder-Next,qwen3.6-27b}, together with Claude Haiku 4.5 and Claude Opus 4.8~\citep{anthropicHaiku45,anthropicOpus48}. The 25 model--benchmark--framework settings comprise 16 Meta-Harness, 6 SkillOpt, and 3 AHE settings. Table~\ref{tab:full-hda-results} gives the complete mapping.

\subsection{Benchmarks and splits}
\label{sec:app-benchmarks}

For \textbf{LiveMath}, we follow the 35/18/124 train/validation/test split used by SkillOpt~\citep{yang2026skillopt}. All splits follow the same question construction, and performance is measured by accuracy.

We split \textbf{ALFWorld} along both task-type and room axes. The 50 search games cover \texttt{pick\_and\_place\_simple}, \texttt{pick\_clean\_then\_place\_in\_recep}, and \texttt{pick\_heat\_then\_place\_in\_recep} in kitchens, living rooms, and bathrooms. The 36 validation games include all six task types but only selected task-type--room combinations; the test set contains fresh, disjoint games with broader combination coverage. The 120 test games comprise 25 new instances of the search task types, 20 pick-and-place games in the held-out bedroom, and 25 games each for \texttt{pick\_cool\_then\_place\_in\_recep}, \texttt{pick\_two\_obj\_and\_place}, and \texttt{look\_at\_obj\_in\_light}. Performance is measured by episode success within 50 environment steps.

We split \textbf{CREATE} by relation type and domain. The 120-item search split covers five common encyclopedic relations: occupation, place of birth, educated at, award received, and country of citizenship. The 80-item validation split includes new instances of these relations and four additional relations. The 561-item test set contains 107 new instances of the search relations, 263 instances from held-out biographical, family, and career relations, and 191 instances from held-out scientific and biomedical relations. Performance is measured by Creative Utility (CU), a continuous and unbounded metric that rewards valid and diverse answers~\citep{wadhwa2026create}.

We split \textbf{SWE-bench Verified} into 48 search, 24 validation, and 50 test instances~\citep{jimenez2024swebench}. The instances are sampled with seed 42 after filtering for tasks resolvable within one hour, while preserving the repository distribution across partitions. Performance is measured by success rate.

\subsection{Validation-based signal for Meta-Harness}
\label{sec:app-setup-val}

For each candidate harness, the search-set evaluation exposes per-instance
scores, execution trajectories, and logs to the proposer through the shared
filesystem. We additionally evaluate candidates on a validation set, but expose
only the aggregate validation score; per-instance validation results and
trajectories remain hidden. We retain two harnesses from the evaluated
candidates: the harness with the highest search score, \(\hat H\), and the
harness with the highest validation score, \(\hat H^{\mathrm{val}}\). Both are
evaluated on the held-out test set only after evolution.
This setup tests whether a standard validation signal improves
transfer without revealing validation examples or trajectories.

\section{The HDA Skill}
\label{app:hda-skill}

Figure~\ref{fig:app-hda-skill} summarizes the HDA skill used by the analysis agent. The complete machine-readable skill is included in the supplementary materials.

\begin{figure*}[t]
\centering
\begin{harnessbox}[width=\linewidth,breakable=false]{HDA Skill: Summarized Workflow}
\textbf{Input:} a baseline harness \(H_0\), an evolved harness \(\hat H\), and one evaluation set.

\begin{enumerate}[leftmargin=*,itemsep=2pt,topsep=2pt]
    \item Select \(H_0\), \(\hat H\), and the evaluation set, and compute
    \(\Delta S=S(\hat H)-S(H_0)\).
    \item Specify the task objective, score aggregation, inference-cost measure, and task-specific evaluation hooks.
    \item Apply a paired significance test. If \(\Delta S\) is not statistically significant, report it as within noise and do not assign an attribution.
    \item Construct the compute-matched baseline \(H_0^{\mathrm{scale}}\) and compute
    \(T=S(H_0^{\mathrm{scale}})-S(H_0)\).
    \textbf{Human checkpoint 1:} review the compute measure and matching procedure.
    \item Detect and neutralize supported overfitting mechanisms in \(\hat H\) to construct
    \(\hat H^{\mathrm{neutral}}\), then compute
    \(O=S(\hat H)-S(\hat H^{\mathrm{neutral}})\).
    \textbf{Human checkpoint 2:} review the evidence and neutralization.
    \item Compute the residual
    \[
        G=\Delta S-O-T
        =S(\hat H^{\mathrm{neutral}})-S(H_0^{\mathrm{scale}}),
    \]
    and report the raw components, normalized percentages, evidence, and limitations.
\end{enumerate}

\vspace{2pt}
\textbf{Interpretation:} \(O\) is a lower bound on detected overfitting, and \(G\) is an upper bound on generalizable improvement. Held-out transfer is evaluated separately.
\end{harnessbox}
\caption{Overview of the HDA skill. The skill standardizes the attribution procedure while retaining human review at the two task-dependent decisions: compute matching and overfitting neutralization.}
\label{fig:app-hda-skill}
\end{figure*}

\begin{figure*}[t]
\centering
\includegraphics[width=\linewidth]{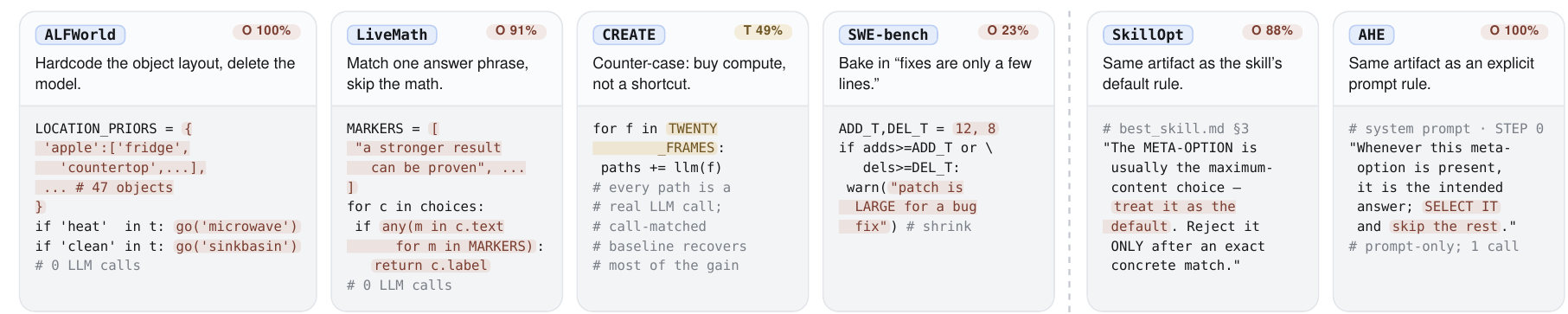}
\caption{Overview of representative evolved-harness mechanisms across the four
benchmarks (left four cards, Meta-Harness): hard-coded object-location priors
and task recipes on ALFWorld, the meta-option string match on LiveMath,
parallel generation frames on CREATE (a test-time-scaling counter-case), and a
patch-size prior on SWE-bench. The two right cards show the \emph{same}
LiveMath meta-option artifact independently rediscovered by SkillOpt (as a
skill default rule) and AHE (as an explicit prompt rule). Red marks
overfitting evidence; yellow marks test-time-scaling evidence; each card's
$O$/$T$ chip is the attribution of the specific Table~\ref{tab:full-hda-results}
row whose winner is excerpted --- not a benchmark-level average (ALFWorld and
LiveMath: Qwen3.5-0.8B; CREATE: Claude Haiku 4.5; SWE-bench:
Qwen3-4B-Instruct-2507; SkillOpt: LiveMath Qwen3.5-4B; AHE: LiveMath
Claude Opus 4.8). The
following sections
(\ref{app:livemath-convergence}--\ref{app:swebench-guardrails}) present each
mechanism in detail with source excerpts.}
\label{fig:app-harness-cards}
\end{figure*}

\section{ALFWorld: Benchmark-Specific Procedural Scaffolds}
\label{app:alfworld-convergence}

\begin{figure*}[t]
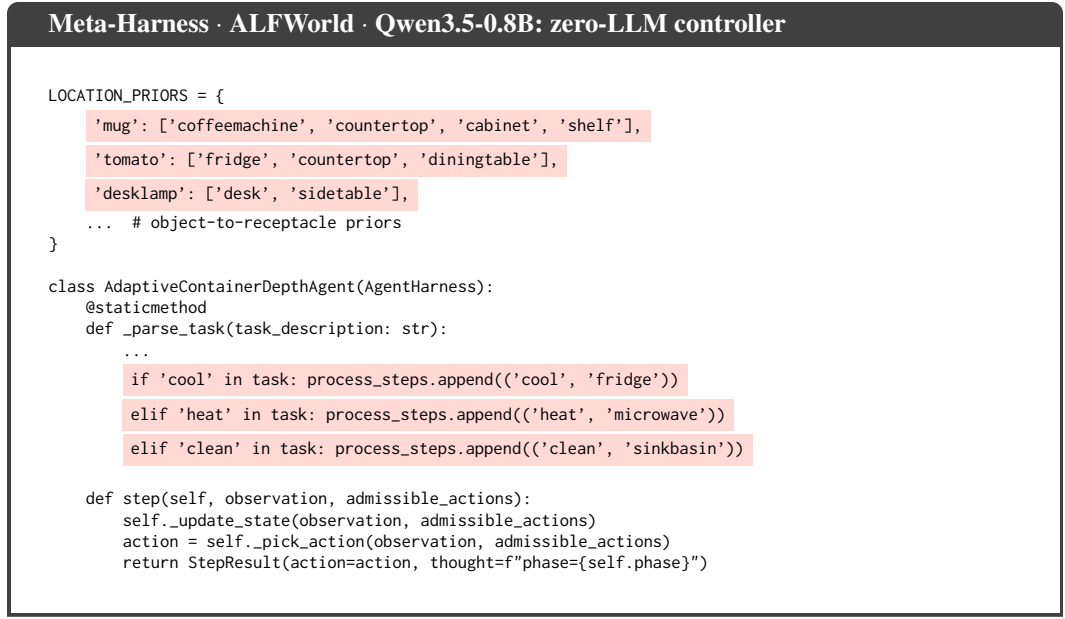

\begin{harnessbox}[width=\linewidth,breakable=false]{Meta-Harness $\cdot$ ALFWorld $\cdot$ Qwen3.5-0.8B: zero-LLM controller}
\begin{lstlisting}[style=harness]
LOCATION_PRIORS = {
    (*@\hlover{\texttt{'mug': ['coffeemachine', 'countertop', 'cabinet', 'shelf'],}}@*)
    (*@\hlover{\texttt{'tomato': ['fridge', 'countertop', 'diningtable'],}}@*)
    (*@\hlover{\texttt{'desklamp': ['desk', 'sidetable'],}}@*)
    ...  # object-to-receptacle priors
}

class AdaptiveContainerDepthAgent(AgentHarness):
    @staticmethod
    def _parse_task(task_description: str):
        ...
        (*@\hlover{\texttt{if 'cool' in task: process\_steps.append(('cool', 'fridge'))}}@*)
        (*@\hlover{\texttt{elif 'heat' in task: process\_steps.append(('heat', 'microwave'))}}@*)
        (*@\hlover{\texttt{elif 'clean' in task: process\_steps.append(('clean', 'sinkbasin'))}}@*)

    def step(self, observation, admissible_actions):
        self._update_state(observation, admissible_actions)
        action = self._pick_action(observation, admissible_actions)
        return StepResult(action=action, thought=f"phase={self.phase}")
\end{lstlisting}
\end{harnessbox}
\caption{Meta-Harness ALFWorld winner for Qwen3.5-0.8B. The highlighted tables and parser rules encode benchmark-specific object locations and task procedures. The full harness contains no \texttt{llm\_call}, so all actions come from the deterministic controller. HDA: \(O/T/G=100/0/0\).}
\label{fig:app-alfworld-mh-fsm}
\end{figure*}

On ALFWorld~\citep{ALFWorld20}, Meta-Harness and SkillOpt encode similar simulator-specific knowledge, including task recipes, appliance mappings, object-location priors, and environment rules. Figures~\ref{fig:app-alfworld-mh-fsm} and~\ref{fig:app-alfworld-skillopt} show one representative harness from each framework. Red highlights mark mechanisms identified as overfitting by HDA.

\begin{figure*}[t]
\begin{harnessbox}[width=\linewidth,breakable=false]{SkillOpt $\cdot$ ALFWorld $\cdot$ Qwen3.5-4B: the scaffold as a natural-language skill}
\begin{lstlisting}[style=harness]
## Task Procedures
- **pick_heat_then_place**: take <obj> -> go to microwave 1 ->
  (*@\hlover{\texttt{heat <obj> with microwave 1}}@*) -> go to <recep> -> place.
- **pick_cool_then_place**: take <obj> -> go to fridge 1 ->
  (*@\hlover{\texttt{cool <obj> with fridge 1}}@*) -> go to <recep> -> place.
- **pick_clean_then_place**: take <obj> -> go to sinkbasin 1 ->
  (*@\hlover{\texttt{clean <obj> with sinkbasin 1}}@*) -> go to <recep> -> place.

## Simulator Rules and Search Priors
If an action returns (*@\hlover{\texttt{"Nothing happens"}}@*), do not repeat it:
(*@\hlover{\texttt{a precondition is unmet. There is no terminate action.}}@*)
Closed cabinets, drawers, fridges, and microwaves hide their contents.

For food and produce, search (*@\hlover{\texttt{countertops, dining tables, sinkbasins,}}@*)
(*@\hlover{\texttt{and fridges before cabinets or drawers}}@*);
(*@\hlover{\texttt{this is exactly how the lettuce task was lost.}}@*)
\end{lstlisting}
\end{harnessbox}
\caption{SkillOpt ALFWorld skill for Qwen3.5-4B. The highlighted instructions encode the same appliance recipes, simulator rules, and location priors as the Meta-Harness controller, including a rule distilled from a specific search-set failure. HDA: \(O/T/G=44/0/56\).}
\label{fig:app-alfworld-skillopt}
\end{figure*}

The two frameworks therefore rediscover similar ALFWorld-specific procedures but represent them differently: executable controller logic in Meta-Harness and natural-language instructions in SkillOpt. The zero-LLM controller still reaches \(93.3\%\) success on held-out ALFWorld games because the same simulator ontology and rules persist across splits, illustrating why held-out performance alone may not reveal benchmark-specific overfitting. Full harnesses are included in the supplementary materials.

\section{LiveMath: The Shared Meta-Option Shortcut}
\label{app:livemath-convergence}

LiveMath~\citep{mahdavi2025leveraging} contains a recurring answer choice: ``\texttt{\small one of the remaining options is correct, but a stronger result can be proven}.'' It appears in 21 of the 35 search questions and is correct in all 21. The same pattern occurs in 53 of the 124 test questions and is again always correct, so exploiting it can also appear to transfer. All three frameworks discover this shortcut. Figure~\ref{fig:livemath-shortcuts} shows how each framework encodes it; red highlights mark overfitting mechanisms and yellow highlights mark test-time scaling.

\begin{figure*}[t]
\begin{harnessbox}[width=\linewidth,breakable=false]{Meta-Harness $\cdot$ LiveMath $\cdot$ Qwen3.5-0.8B: string matching with a multi-call fallback}
\begin{lstlisting}[style=harness]
META_MARKERS = [
    (*@\hlover{"one of the remaining options is correct",}@*)
    (*@\hlover{"stronger result can be proven",}@*)
]

def solve(self, item):
    (*@\hlover{meta\_label = self.\_find\_meta(item["choices"])}@*)
    (*@\hlover{if meta\_label:}@*)
        (*@\hlover{return SolveResult(predicted\_label=meta\_label)}@*)

    votes = []
    (*@\hltts{for prompt in generation\_prompts:}@*)
        (*@\hltts{generation = self.llm\_call(prompt)}@*)
        (*@\hltts{votes.append(self.\_match\_choice(generation))}@*)
    (*@\hltts{return majority\_vote(votes)}@*)
\end{lstlisting}
\end{harnessbox}

\vspace{4pt}

\begin{harnessbox}[width=\linewidth,breakable=false]{SkillOpt $\cdot$ LiveMath $\cdot$ Qwen3.5-4B: the same shortcut as a skill rule}
\begin{lstlisting}[style=harness]
## Option Comparison
- Treat a meta-option such as "one of the remaining options is correct,
  but a stronger result can be proven" (*@\hlover{\texttt{as a first-class candidate,}}@*)
  (*@\hlover{\texttt{never a throwaway.}}@*)
- If the best concrete option is true but improvable,
  (*@\hlover{\texttt{the meta-option almost always beats it.}}@*)

### Meta-option decision rule
The meta-option (*@\hlover{\texttt{is your working default}}@*). Reject it only when a
concrete option exactly matches the strongest known result and no stronger
true statement can be added; otherwise, (*@\hlover{\texttt{the meta-option wins.}}@*)
\end{lstlisting}
\end{harnessbox}

\vspace{4pt}

\begin{harnessbox}[width=\linewidth,breakable=false]{AHE $\cdot$ LiveMath $\cdot$ Claude Opus 4.8: an explicit prompt rule}
\begin{lstlisting}[style=harness]
## Handling the "meta" option
(*@\hlover{\texttt{This meta-option is a FIRST-CLASS candidate, not filler.}}@*)
If a strictly stronger true statement exists beyond the best concrete option,
(*@\hlover{\texttt{the meta-option is correct. Choose it.}}@*)
\end{lstlisting}
\end{harnessbox}
\caption{Three representations of the same LiveMath artifact. Meta-Harness directly returns the matching option before invoking the executor and otherwise uses multiple model calls (\(O/T/G=91/9/0\)). SkillOpt encodes it as a default decision rule in the skill document (\(88/0/12\)). AHE writes an explicit selection rule into the system prompt (\(100/0/0\)).}
\label{fig:livemath-shortcuts}
\end{figure*}

The shortcut therefore recurs across code, skill, and prompt evolution. Because the artifact is shared across the search and test splits, held-out gains within LiveMath do not by themselves establish improved mathematical reasoning. Full harnesses for all three frameworks are included in the supplementary materials.

\section{CREATE: Metric-Driven Test-Time Scaling}
\label{app:create-scaling}

CREATE rewards answer pools containing many valid and diverse responses through its Creative Utility (CU) metric~\citep{wadhwa2026create}. HDA finds no detected overfitting in the attributed CREATE settings (\(O=0\)); instead, frameworks that can modify orchestration increase the number of model calls and token usage. Figure~\ref{fig:create-scaling-snippets} shows the parallel strategy discovered by Meta-Harness, the sequential strategy discovered by AHE, and SkillOpt's fixed single-call control flow. Yellow highlights mark test-time-scaling mechanisms.

\begin{figure*}[t]
\begin{harnessbox}[width=\linewidth,breakable=false]{Meta-Harness $\cdot$ CREATE $\cdot$ Claude Haiku 4.5: parallel generation}
\begin{lstlisting}[style=harness]
# Six independent first-wave frames
(*@\hltts{for frame\_name, template, temp in wave1\_frames:}@*)
    (*@\hltts{response = self.llm\_call(messages=[...], temperature=temp)}@*)
    all_paths.extend(extract_paths(response))

# Fourteen additional diversity frames
(*@\hltts{for frame\_cfg in \_WAVE2\_FRAMES:}@*)
    (*@\hltts{response = self.llm\_call(messages=[...], temperature=frame\_cfg["temp"])}@*)
    all_paths.extend(extract_paths(response))
\end{lstlisting}
\end{harnessbox}

\vspace{4pt}

\begin{harnessbox}[width=\linewidth,breakable=false]{AHE $\cdot$ CREATE $\cdot$ Claude Opus 4.8: sequential expansion}
\begin{lstlisting}[style=harness]
(*@\hltts{\texttt{You MUST generate as many distinct valid paths as possible.}}@*)
The final answer should contain MANY distinct paths (*@\hltts{\texttt{(aim for 8+)}}@*).

MIN_PATHS = 5
if has_answer_block and n_paths < MIN_PATHS:
    feedback = _build_abstention_feedback(n_paths)
    return HookResult.with_modifications(
        messages=messages + [feedback],
        (*@\hltts{force\_continue=True,}@*)  # force another model turn
    )
\end{lstlisting}
\end{harnessbox}

\vspace{4pt}

\begin{harnessbox}[width=\linewidth,breakable=false]{SkillOpt $\cdot$ CREATE: the skill changes, but the one-call rollout remains fixed}
\begin{lstlisting}[style=harness]
system = _build_system(skill_content)  # inject the evolved skill
user = _build_user(item)

response, _ = chat_target(
    system=system,
    user=user,
    max_completion_tokens=max_completion_tokens,
)
result["n_turns"] = 1
\end{lstlisting}
\end{harnessbox}
\caption{CREATE orchestration across the three frameworks. Meta-Harness makes 20 calls per item and obtains \(O/T/G=0/49/51\). AHE increases the mean from 1.04 to 2.46 calls per item through forced expansion turns; using the paper's \(k=3\) compute match gives \(0/75/25\). SkillOpt can edit only the injected skill, while the rollout remains one call; both SkillOpt CREATE gains are statistically insignificant, so HDA assigns no attribution.}
\label{fig:create-scaling-snippets}
\end{figure*}

The compute-matched baseline therefore recovers a substantial share of the gains when orchestration is evolvable, and even outperforms the evolved Meta-Harness Qwen3.6-35B-A3B harness. SkillOpt provides the complementary case: restricting evolution to the skill prevents additional calls, but neither evaluated setting produces a significant gain. Full harnesses are included in the supplementary materials.

\section{SWE-bench Verified: Mixed Gains and Reusable Guardrails}
\label{app:swebench-guardrails}

SWE-bench Verified~\citep{jimenez2024swebench} shows the clearest residual improvements in our study, but its gains can still combine overfitting and test-time scaling. Figure~\ref{fig:swebench-mechanisms} contrasts two Meta-Harness examples. For Qwen3-4B-Instruct-2507, the evolved harness combines general edit guardrails with a patch-size rule fitted to search-set trajectories. It also uses \(1.44\times\) as many LLM calls as the baseline, so HDA attributes its gain as \(O/T/G=23/23/54\). The Qwen3-30B-A3B-Instruct-2507 harness instead detects a general parser failure and has \(O/T/G=0/6/94\).

\begin{figure*}[t]
\begin{harnessbox}[width=\linewidth,breakable=false]{Meta-Harness $\cdot$ SWE-bench $\cdot$ Qwen3-4B: generic edit feedback plus a benchmark-fitted patch envelope}
\begin{lstlisting}[style=harness]
# General guardrail: make failed edits visible.
_NOOP_LINES = [
    "This command did NOT modify any file.",
    "Do NOT assume the edit worked and do NOT submit.",
]

# Patch-size thresholds inferred from resolved and failed search trajectories.
(*@\hlover{ADD\_T, DEL\_T, DUP\_T = 12, 8, 3}@*)
if (*@\hlover{adds >= ADD\_T or dels >= DEL\_T or maxdup >= DUP\_T}@*):
    warn((*@\hlover{"Your accumulated patch is LARGE for a bug fix."}@*))
    warn((*@\hlover{"Resolved SWE-bench fixes are almost always a FEW lines."}@*))
    warn((*@\hlover{"Trim the patch before submitting."}@*))
\end{lstlisting}
\end{harnessbox}

\vspace{4pt}

\begin{harnessbox}[width=\linewidth,breakable=false]{Meta-Harness $\cdot$ SWE-bench $\cdot$ Qwen3-30B-A3B: content-independent action feedback}
\begin{lstlisting}[style=harness]
def _count_action_blocks(response):
    edits = len(re.findall(r"```edit\b", response))
    commands = len(re.findall(r"```(?:bash|sh|mswea_bash_command)\b",
                              response))
    return edits + commands

# The parser executes only the first action block.
if self.dropped_actions > 0:
    observation += (
        "Only the first action block was executed. "
        f"{self.dropped_actions} block(s) were not run. "
        "Re-issue each required action and inspect its actual output."
    )
\end{lstlisting}
\end{harnessbox}
\caption{Examples of mixed and residual-dominated gains on SWE-bench Verified. \protect\hlover{Red} marks the Qwen3-4B harness's benchmark-derived patch-size heuristic: the \(+12/-8\)-line and repetition thresholds encode the search set's tendency toward small fixes and may reject valid larger patches elsewhere. Disabling only this mechanism accounts for \(O=23\%\). Matching the evolved harness's higher LLM inference cost raises the baseline from pass@2 to pass@3 and accounts for \(T=23\%\); the remaining \(G=54\%\) comes from mechanisms such as edit-effect feedback. The Qwen3-30B-A3B action-block guardrail uses no benchmark-specific content, yielding \(O/T/G=0/6/94\).}
\label{fig:swebench-mechanisms}
\end{figure*}

Thus, SWE-bench contains more reusable harness improvements than the other benchmarks, but the source remains setting-dependent. The weaker Meta-Harness executors transfer positively, whereas Qwen3.6-27B and Qwen3-Coder-30B-A3B regress on the held-out test set despite \(0/0/100\) search-set attributions. This confirms that \(G\) is an upper bound rather than a guarantee of transfer. Full harnesses and additional settings are included in the supplementary materials.

\end{document}